\documentclass{article}

\usepackage[a4paper]{geometry}                
\usepackage{epsfig}
\usepackage{graphicx}
\usepackage{amsmath,amsthm}
\usepackage{amssymb}
\usepackage{algorithmic}
\usepackage{algorithm}
\usepackage{subfigure}
\usepackage{hyperref}
\usepackage[applemac]{inputenc}
\usepackage{pst-sigsys}
\graphicspath{{../figures/}}

\date{\ }                                           

\newtheorem{theorem}{Theorem}[section]
\newtheorem{lemma}[theorem]{Lemma}

\newtheorem{proposition}[theorem]{Proposition}

\title{Deep Learning by Scattering}

\author{St\'ephane Mallat and Ir\`ene Waldspurger\\
Depart. d'Informatique, \'Ecole Normale Sup\'erieure\\
45 rue d'Ulm, Paris, France}

\newcommand{\m}{{\overline m}}

\newcommand{\E}{{\bf E}}

\newcommand{\R}{{\mathbb{R}}}
\newcommand{\C}{{\mathbb{C}}}

\newcommand{\Z}{{\mathbb{Z}}}

\newcommand{\N}{{\mathbb{N}}}

\newcommand{\rb}{{\rangle}}
\newcommand{\lb}{{\langle}}

\newcommand{\X}{{X}}
\newcommand{\tX}{{\overline X}}
\newcommand{\tY}{{\overline Y}}
\newcommand{\tU}{{\overline U}}
\newcommand{\tx}{{\bar x}}
\newcommand{\ty}{{\bar  y}}

\newcommand{\Y}{{ Y}}
\newcommand{\U}{{ U}}

\begin{document}
\maketitle
~\\
\begin{abstract}
We introduce general
scattering transforms as mathematical models of deep neural 
networks with $\bf l^2$ pooling.
Scattering networks iteratively apply complex valued 
unitary operators, and the pooling is performed by a complex modulus.
An expected scattering defines a contractive representation 
of a high-dimensional probability distribution, which preserves its
mean-square norm. We show that
unsupervised learning can be casted as an optimization of 
the space contraction to
preserve the volume occupied by unlabeled examples, at each layer of the
network. Supervised learning and
classification are performed with an averaged scattering, which provides
scattering estimations for multiple classes.
\end{abstract}

\section{Introduction}

Hybrid generative and discriminative classifiers are powerful
when there is a large databases of unlabeled examples and a much smaller
set of labeled examples \cite{Hybrid}. Building such classifiers requires to address two
outstanding problems: estimating and representing a high dimensional
probability distribution from unlabeled examples and integrating this
representation in a supervised classifier. 

Deep neural networks are remarkable
implementations of this strategy, which has produced state-of-the-art results
in many fields including image, video, music,
speech and bio-medical data \cite{Bengio}. 
Most deep neural networks cascade linear operators followed
by ``pooling non-linearities'' which aggregate multiple variables
\cite{LeCun,Hinton,Le,Ranzato,Bengio2}. 
The hidden network variables are first estimated by unsupervised learning
and are then updated together with the optimization of a supervised classifier
from labeled examples. Multiple regularization criteria such as
contraction and sparsity have been shown to play an important role in
deep networks \cite{Bengio2}. Despite 
the multiplicity of architectures, algorithms and results,
there is currently a lack of mathematical models
to understand their behavior.

This paper introduces a mathematical and algorithmic 
framework, to analyze the properties of high dimensional
unsupervised and supervised classification problems with deep networks. 
It is built on a general scattering model of deep networks
with $\bf l^2$ pooling, which
iterates on contraction operators obtained as the complex modulus of
unitary linear operators. It relies on many ideas developed in the
deep network literature \cite{Bengio}.
Section \ref{expnsdfs} proves that an expected scattering
transform defines a converging deep network whose output defines a
representation of the underlying probability distribution.
Section \ref{option} shows that such representations are optimized
by adaptively contracting the space while preserving the
volume where the distribution of unlabeled examples is concentrated.
The optimization establishes a relation with sparsity, 
which explains mathematically why 
sparse regularizations are efficient for deep network learning \cite{Bengio}.
Section \ref{Ergodic} explains how scattering models can
be estimated from a single realization with averaged scattering transforms,
whose properties are analyzed. Given the large body of numerical experiments
in the deep network literature, the paper concentrates on mathematical models,
algorithms and proofs, which are currently lacking \cite{Bengio,Hinton,Ranzato,LeCun}.

{\bf Notation:} 
The modulus of $z = a + i b \in \C$ is written $|z| = (|a|^2 + |b|^2)^{1/2}$.
If $x = (x_n)_{n \leq N} \in \C^N$ then we write $|x| = (|x_n|)_{n \leq N} \in \R^N$.

\section{Expected Scattering}
\label{expnsdfs}
A scattering transform provides 
a model for feed-forward deep networks with $\bf l^2$ pooling 
\cite{Le,Ranzato}. 
It iterates on linear unitary operators
followed by complex modulus. 
Scattering transforms have initially been introduced with wavelet
operators to build invariants to translations, which are
stable to deformations, with applications
to image and audio classifications \cite{Anden,Joan}. 
The following generalization only imposes the use of unitary operators, that
we shall optimize from examples. It covers both convolutional
and non-convolutional deep networks. 

Let $X$ be a random vector defined in $\R^N$. 
We initialize $\X_0  = X$ and $N_0 = N$.
An expected scattering 
computes each network layer $\X_{m+1} \in \R^{N_{m}}$ 
by transforming the previous layer $\X_m$ with an operator
$W_{m+1} $ from $\R^{N_{m}}$ in $\C^{N_{m+1}}$ such that
\[
W_m^*\, W_m = Id~. 
\]
We typically have $N_{m+1} > N_m$ so $W_m$ is represented by a complex valued
matrix whose rows are linearly dependent complex vectors. With an abuse of
language, we still say that these operators are unitary.
The $\bf l^2$ pooling is 
implemented with a complex modulus along each coordinate:
\begin{equation}
\label{intsdf}
\X_{m+1}  = |W_{m+1} (\X _{m}  -  E (\X_{m} ))|~.
\end{equation}
Let $UX = \{ X_m\}_{m \in \N}$ be
the set of all propagated layers.
An expected scattering transform outputs
\begin{equation}
\label{scasdfns}
E(U X) = \{ E(\X_m ) \}_{m \in \N} ~. 
\end{equation}
This provides a representation of the probability distribution of $X$.

The operators $W_{m+1}$ encode the weights of the feed-forward network. 
Each unitary operator can be written
$W_{m+1} x = \Big\{ \lb x , \psi_n \rb \Big\}_{n \leq N_{m+1}}$ with 
$\psi_n = \psi_n^a + i \psi_n^b\in \C^{N_{m}}$, where
$\{ \psi_n \}_{n \leq N_m}$ is a tight frame of $\R^{N_{m}}$. 
It groups pairs of random variables
$\lb \X_m -E(\X_m), \psi^a_n \rb$ and 
$\lb \X_m -E(\X_m), \psi^b_n \rb$, whose variabilities are
reduced by the contractive aggregation of the complex modulus
\[
|\lb \X_m -E(\X_m), \psi_n \rb| = 
\Big(|\lb \X_m -E(\X_m), \psi^a_n \rb|^2 
+ |\lb \X_m -E(\X_m), \psi^b_n \rb|^2\Big)^{1/2}~. 
\]
Ideally, $W_{m+1}$ groups pairs of non-correlated
random variables having the same variance, so that it 
reduces the process variability without suppressing correlation information.
A scattering can also 
pool and compute the $\bf l^2$ norm of $2^k$ variables instead of just $2$,
by cascading each time $k-1$ more scattering
contractions. This pooling is defined 
by operators $W_m$ which aggregates variables
by pairs $(k',k'')$ with  $\psi_n(k) = \delta_{k'}(k) + i \delta_{k''}(k)$,
where $\delta_{k'} (k) = 1$ if $k=k'$ and $0$ otherwise,
to progressively build each pool of $2^k$ variables.

The operator $W_{m+1}$ typically performs a rotation of the space
to optimize the contraction, which is related to sparsity,
as explained in the next section. Redundancy and hence increasing the
space dimension is important to improve
sparsity. Redundancy also prevents 
loosing information when calculating the modulus. 
With a redundancy factor $4$,
one can indeed build operators $W_{m+1}$ from
$\R^N$ to $\C^{2N}$ such that any $x$ has a stable 
recovery from $|W_{m+1} x|$ \cite{Candes}. 
The following theorem proves that for any set of unitary operators,
the expected scattering transform
is contractive and preserves energy. The proof is in 
Appendix \ref{blabla}.

\begin{theorem}
\label{theonsdfsdf}
The scattering operator is contractive
\begin{equation}
\label{contsf}
\|E(U X) - E(U Y) \|^2 = \sum_{m=0}^{\infty} \|  E(\X_m ) -  E(\Y_m ) \|^2
\leq E (\|X - Y \|^2)~,
\end{equation}
and preserves the mean-square norm
\begin{equation}
\label{nsdfoins}
\|E(U X) \|^2 = \sum_{m=0}^\infty  \|E(\X_{m}) \|^2 = E(\|X\|^2) ~.
\end{equation}
If $\|X\|$ is not bounded with probability $1$ then 
\begin{equation}
\label{nsdfoins2}
\sum_{m=0}^{\infty} \|E(\X_m)\| = \infty~
\end{equation}
\end{theorem}

The theorem proof shows that the energy 
of network layers $E(\|\X_m \|^2)$ converges to $0$ as $m$ increases
despite the fact that the dimension $N_m$ of these layers may increase
to $\infty$. The convergence is exponential at first and then goes into
a slow decay asymptotic regime, which explains why $\|E(\X_m)\|$ is not 
summable.
Expected scattering transforms
computed with wavelet transforms operators $W_{m}$ define deep convolution
networks \cite{LeCun}, which are highly effective for a number of
image and audio classification problems
\cite{Anden,Joan}. The operator $W_1$ decomposes $x \in \R^N$ into
$J$ complex wavelet signals of size $N$, so $N_1 = J  N$. 
The operator $W_2$ transforms the modulus of 
each of these wavelet signals into yet again
$J$ wavelet signals of size $N$, so $N_2 = J^2 N$. The $m^{th}$ wavelet
layer is thus of size $N_m = J^m N$ which grows 
exponentially to $\infty$ when $m$ increases. 
For processes which are not bounded, (\ref{nsdfoins2}) proves that the
decay of scattering coefficients is asymptotically very slow. During the
first iterations $\|E(\X_m)\|$ decays exponentially but it then slows down
and decay slowly. In this slow regime, scattering coefficients characterize
the tail of the probability distribution.

An expected scattering transform specifies a unique probability distribution
of maximum entropy. Let us write $\U_m X = \X_m$.
Given an expected scattering transform $E(U X) = \{ E(U_m X) \}_{m \in \Z}$,
the Boltzmann theorem proves that the
probability density $\tilde p(x)$ of maximum entropy which satisfies
\begin{equation}
\label{constra}
\forall m \in \N~~,~~
\int_{\R^{N}}\,  \U_m x\, \tilde p(x)\, dx =  E(\U_m X)  
\end{equation}
can be written
\begin{equation}
\label{constra2}
\tilde p(x) = \frac 1 {Z} \exp\left(- \sum_{m=0}^{\infty}  \lb \lambda_m ,  \U_m x \rb
\right)~
\end{equation}
where the $\lambda_m \in \R^{N_m}$ are the 
Lagrange multipliers associated to the constraints (\ref{constra}), and 
$Z$ is the normalization partition function.
Complex audio textures are efficiently synthesized 
with such models \cite{Joan,Anden}, by using wavelet
operators $W_m$. 

\section{Unsupervised Learning by Optimizing Contractions}
\label{option}

A scattering transform progressively squeezes the
space. Heuristics are most often used to regularize
unsupervised optimizations \cite{bengio}.
Because all operators are unitary, we show that
optimizing this contraction 
amounts to minimizing the decay of scattering coefficients and 
leads to sparse representations. Estimators of
expected scattering coefficients are given with error bounds.

Unsupervised learning considers a mixture $X$ of unknown classes
$\{X^{(k)} \}_k$. We want to optimize each $W_m$ to then be able to
discriminate each mixture component (class) at the supervised
classification stage. To avoid confusing the scattering representations
$E(U X^{(k)})$ of different classes, 
we would like to find operators $W_m$ which maximizes the average
scattering distance between classes:
\begin{equation}
\label{averadisngs0df}
\sum_{k,l} p_k \, p_l \|E(U X^{(k)}) - E(U X^{(l)})\|^2~,
\end{equation}
where $p_k$ is the probability of $X^{(k)}$ in
the mixture $X$.
However, unsupervised learning cannot minimize this average distance 
since we do not know the class labels $k$.

Following the greedy layerwise unsupervised learning strategy introduced by
Hinton \cite{Hinton}, we build the scattering transform layers one after the
other, for increasing depth. 
We thus suppose that all operators $W_n$ are
defined for $n \leq m$, before optimizing $W_{m+1}$. 
Since the average distance (\ref{averadisngs0df}) of
mixture components cannot be computed, it is 
replaced by a maximization of the mixture variance $E(\|X_{m+1} - E(X_{m+1})\|^2)$.
Since $W_{m+1}$ is unitary, and
$\X_{m+1}  = |W_{m+1} (\X _{m}  -  E (\X_{m} ))|$, we derive that
\[
E (\|\X_{m} - E(\X_{m}) \|^2) - 
E (\|\X_{m+1}- E(X_{m+1})\|^2)  = 
\|E( \X_{m+1}) \|^2 ~. 
\]
Given $\X_{m}$, maximizing the variance of $\X_{m+1}$ is equivalent to find 
a unitary operator $W_{m+1}$ which minimizes 
\begin{equation}
\label{ndsfu89s0}
\|E( \X_{m+1}) \|^2 = \| E( |W_{m+1} (\X_{m} - E(\X_{m}))|) \|^2~.
\end{equation}
Minimizing the norm of this expected value enforces the sparsity of coefficients
across realizations. It creates a deep network which
filters realizations of $X$ so that their energy propagates across the
network, as opposed to other signals which are not sparsified by the
operators $W_{m+1}$ and will thus be attenuated much faster.

Expected scattering coefficients are estimated
from $P$ independent examples $\{X_i \}_{i \leq P}$.
A scattering transform of each $X_i$ is calculated by 
initializing $\overline X_{i,0} = X_i$. For each $m \geq 0$,
given $\{ \overline X_{i,m} \}_{i \leq P}$ we compute an
estimator of $E(\X_{m})$ with an empirical average
\begin{equation}
\label{dsfoinsd58}
\overline \mu_{m} = P^{-1} \sum_{i=1}^P \overline X_{i,m}~.
\end{equation}
The scattering iteration replaces
$E(\X_{m})$ by $\overline \mu_{m}$ in (\ref{intsdf}) which defines
\begin{equation}
\label{dsfoinsd58}
\overline X_{i,m+1} = |W_{m+1} (\overline X_{i,m} - \overline \mu_{m} ) |~.
\end{equation}
Section \ref{Ergodic} generalizes this
estimated scattering by introducing an averaged scattering operator.
Appendix \ref{blabla2} proves the following upper bound of the 
mean-square estimation error.

\begin{theorem}
\label{firnsdfpors}
\begin{equation}
\label{dsfoinsd5}
E(\| \overline \mu_m - E(\X_m)\|^2) 
\leq P^{-1} \left(\sum_{n=0}^m\left(\sum_{k=n+1}^{\infty} \|E(\X_k)\|^2\right)^{1/2}
\right)^2 \leq P^{-1} (m+1)^2 E(\|X\|^2)~.
\end{equation}
\end{theorem}

Numerically, the first upper bound is 
typically of the order of $P^{-1} E(\|X\|^2)$. The estimation error is
therefore small relatively to $\|E(X_m)\|^2$ if
$P \gg E(\|X\|^2) / \|E(\X_m)\|^2$.

To optimize the operator
$W_{m+1} x = \{ \lb x , \psi_n \rb \}_{n}$ we estimate
(\ref{ndsfu89s0}) with a summation across examples:
\begin{equation}
\label{ndsfu89s}
P^{-2} \Big\| \sum_{i \leq P} |W_{m+1} (\overline{X}_{i,m} - \overline \mu_{i,m})| \Big\|^2 =
P^{-2} \sum_{n=1}^{N_{m+1}}\left( \sum_{i \leq P} \Big
|\lb \overline{X}_{i,m} - \overline \mu_{i,m} , \psi_n \rb \Big|\right)^2~.
\end{equation}
Minimization of such a
convex functional under the unitary condition $W_{m+1}^* W_{m+1} = Id$ 
is a Procrustes type optimization, which
admits a convex 
relaxation formulation as a Semi Definite Positive optimization
\cite{SDP}. Nearly optimal solutions can thus be computed, although the
resolution of these SDP problems are numerically expensive. Stochastic
gradient descent algorithms are typically used in applications \cite{Le}.

Through this minimization, the complex modulus in (\ref{ndsfu89s})
tends to define $\psi_n = \psi_n^a + i \psi_n^b$ which groups
non-correlated random variables
$\lb \tX_{i,m} - \overline \mu_{i,m} , \psi^a_n \rb$ and
$\lb \tX_{i,m} - \overline \mu_{i,m} , \psi^b_n \rb$ with
same variance. The summation over $i$ defines an
$\bf l^1$ norm across different realizations.
Minimizing this norm
enforces the sparsity of this
sequence. It produces few large coefficients and many small ones.
The sparsity across realizations $i$ also implies a sparsity across $n$
for most realizations because the overall family of coefficients
$\{ |\lb \tX_{i,m} - \overline \mu_{i,m} , \psi_n \rb|\}_{i,n}$ is sparse.

\section{Averaged Scattering}
\label{Ergodic}

We now explain how to compute hybrid generative and discriminative
classifiers from scattering transforms, which integrate unsupervised
learning with a supervised classification. It gives a mathematical model
to explain the supervised refined training of deep neural networks, from
an initial unsupervised training. 
Section \ref{option} explains how to use unlabeled
examples to optimize the unitary operators $W_m$, in order to preserve
the discriminability property of 
the expected scattering transform. Given few labeled
examples for each class $X^{(k)}$, (\ref{dsfoinsd58}) computes an estimation
of $E(U X^{(k)})$ whose risk is bounded by (\ref{dsfoinsd5}). 
To classify a signal $x$, which is the
realization of any unknown class $X^{(l)}$,
we introduce an averaged scattering transform, which provides
an estimator of $E(U X^{(l)})$.

An averaged scattering transform of a vector $x \in \R^N$ is initialized with
$\tx_0 = x$. Each expected value is estimated by a
block averaging $A_{m}$ applied to the network layer $\tx_{m} \in \R^{N_{m}}$.
It averages $\tx_m$ over blocks $B_{j,m}$ of size $B_{j,m}^\#$,
which define a partition of $\{1,...,N_{m}\}$:
\[
A_{m} \tx_{m} (n) = \sum_j \frac 1 {B_{j,m}^{\#}}\left(\sum_{k \in B_{j,m}} \tx_m (k)
\right)\, 1_{n \in B_{j,m}}~.
\]
The next layer of scattering coefficients is computed by applying the unitary
operator $W_{m+1}$:
\begin{equation}
\label{intsdf22}
\tx_{m+1} = |W_{m+1} (\tx_{m} - A_{m} \tx_{m})|= |W_{m+1} \,(Id-A_{m}) \tx_{m}| ~.
\end{equation}
The averaged scattering transform outputs
the block averages of all layers $\tU x = \{ \tx_m \}_{m \in \N}$:  
\begin{equation}
\label{scasdfns2}
A \tU x = \{ A_{m} \tx_m  \}_{m \in \N} ~
\end{equation}

\begin{theorem}
\label{condsfth}
The averaged scattering operator is contractive
\begin{equation}
\label{contsf9}
\|x - y \|^2 \geq \|A \tU x - A \tU y \|^2 = \sum_{m=0}^{\infty} \|A_{m} \tx_m - A_{m} \ty_m \|^2~.
\end{equation}
If $A_m$ averages over blocks of size at most $M$ then $\|\tx_m \|^2 \leq \|x\|^2\, (1-M^{-1})^{m}$ and
\begin{equation}
\label{nsdfoins3}
\|x\|^2 = \|A \tU x \|^2 = \sum_{m=0}^\infty  \|A_m \tx_m\|^2 ~.
\end{equation}
\end{theorem}

The unitary operators $W_m$ are optimized at the unsupervised stage,
by maximizing a variance criterion which tends to increase
the average distance between the unknown
scattering vectors $E(U X^{(k)})$ of each class. At the supervised
stage each $E(U X^{(k)})$ is estimated with an error bounded by
Theorem \ref{firnsdfpors}. A generative classifier needs to 
optimize the block averages $A_m$ so that 
$A \tU X^{(k)} = \{ A_m \tX^{(k)}_m \}_m$ gives an accurate estimator of 
$E(U X^{(k)}) = \{E(\X_m)\}_m$. As a result, the 
class of a signal $x$ can simply be estimated by
\[
\hat k = \arg \min_{k} \|A \tU x - E(U X^{(k)}) \|~.
\]
The error rate of such a classifier depends upon the 
estimation error of $E(U X^{(k)}) = \{E(\X^{(k)}_m) \}_{m \in \N}$ 
by $A \tU x = \{A_m \tx_m \}_{m \in \N}$ when $x$ is a realization of $X^{(k)}$. 
The following proposition computes an upper bound on this error.

\begin{figure}
\begin{center}
\includegraphics[width=0.8\columnwidth]{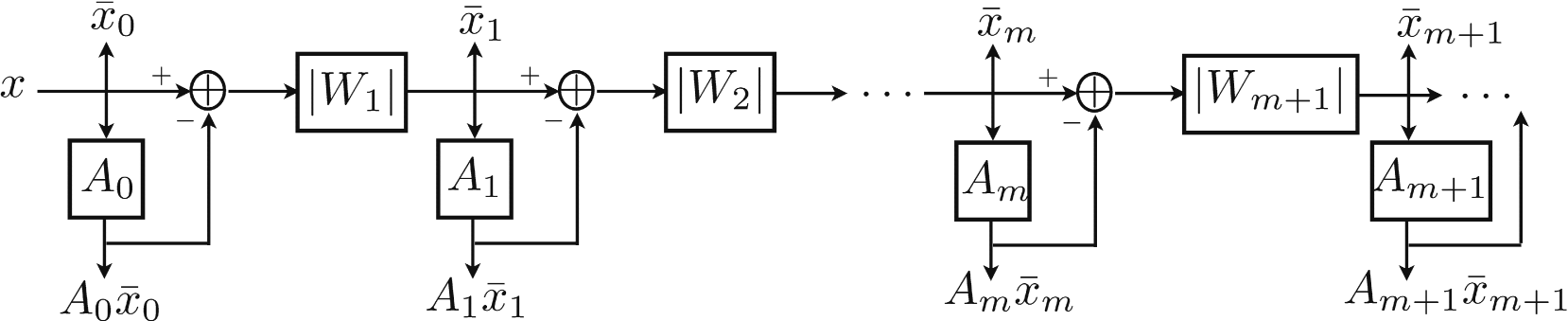}
\end{center}
\caption{An averaged scattering network iteratively computes 
$\tx_{m+1} = |W_{m+1} (\tx_m - A_m \tx_m)|$. It outputs
$\tU =\{ \tx_m \}_m$ or $A \tU = \{A_m \tx_m \}_m$ for classification.}
\label{averagsndf}
\end{figure}

\begin{proposition}
\label{ensdfonfops}
\begin{equation}
\label{dsfoinsd3}
E(\|A_m \tX^{(k)}_{m} - E(\X^{(k)}_{m})\|^2) \leq 
\left(\sum_{n=0}^m E(\|A_n \X^{(k)}_{n} - E(\X^{(k)}_{n})\|^2)^{1/2} \right)^2 ~.
\end{equation}
\end{proposition}

The estimation error $E(\|A_m \tX^{(k)}_{m} - E(\X^{(k)}_{m})\|^2)$
is small if the averaging bias $\|A_m E(\X^{(k)}_m) - E(\X^{(k)}_m) \|^2$ 
and the variance $E(\|A_m \X^{(k)}_m - A_m E(\X^{(k)}_m) \|^2)$ are small for all
$m$ and $k$. It means that $A_m$ should average the largest possible
groups of coefficients where $E(\X^{(k)}_m)$ has a small variation for all $k$.
Prior information is usually available to constrain the $A_m$. 
In sounds or images for example, 
the averaging is partly done in time or space (but not only),
over intervals that must be adjusted according to the unknown
local stationarity property of the $\X^{(k)}$. This is the case for audio
and image texture discrimination with wavelet
scattering networks \cite{Joan,Anden}.

Discriminative classifiers typically outperform generative classifiers,
and be computed directly 
from $\tU x$ as opposed to $A \tU x$. Let us consider a binary
linear classifier such as an SVM, which applies classification thresholds 
to $\lb w , A \tU x \rb$ for an optimized vector $w$.
Since $A$ is a linear projector, 
$\lb w , A \tU x \rb = \lb w' , \tU x \rb$ with $w' = A w$. 
The linear classification can thus be directly applied on $\tU x$. Results
may be improved by using $A \tU x$ only if 
$A$ is a way to incorporate prior information such as local stationarity properties. 

Since $\tU x = \{ \tx_m \}_m$ 
is computed in (\ref{intsdf22}) by iterating on 
 $W_{m} (Id - A_{m-1})$, the supervised classifier still
needs to optimize the choice of the $A_m$. It amounts to replacing
the $W_m$ calculated by unsupervised
learning by a new operator 
$W_m (Id - A_{m-1})$ which is optimized to minimize
the classification error. 
Deep neural networks perform such an update of the network parameters,
with a greedy layerwise supervised optimizations of the neuron weights
\cite{Bengio}. This last step depends upon the type of discriminative
classifier which is used.

\section{Conclusion}

A scattering transform provides a flexible model for general deep networks
with $\bf l^2$ pooling.
Imposing that linear operators are unitary preserves
information and stability, and defines a network whose properties can be analyzed mathematically. It provides new models for high-dimensional probability 
distributions, with precise bounds on estimation errors from samples.
Network parameters are optimized from unlabeled examples by 
adjusting the space contraction, which admits an SDP convex relaxation.
Supervised classifiers are computed with an averaged scattering, which
is initialized by the unsupervised estimation and refined from labeled
examples.

\appendix
\section{ Proof of Theorem \protect \ref{theonsdfsdf}}
\label{blabla}

Observe that 
$|W_m|$ is a contractive operator because $W_m$ is unitary.
So
\begin{eqnarray}
\label{sdfsdf}
E(\|\X_m - \Y_m\|^2) &\leq&
E(\|\X_{m-1} - \Y_{m-1} -  E(\X_{m-1} - \Y_{m-1})\|^2) \\
\nonumber
&=&E(\|\X_{m-1} - \Y_{m-1}\|^2)  - \| E(\X_{m-1} - \Y_{m-1}) \|^2 .
\end{eqnarray}
It results that
\[
E(\|\X_{0} - \Y_{0}\|^2)  \geq \sum_{m=1}^{\m} 
\| E(\X_{m-1} - \Y_{m-1}) \|^2 + E(\|\X_{\m} - \Y_{\m}\|^2) ~.
\]
Letting $\m$ go to $\infty$ proves (\ref{contsf}).
If we set $Y = 0$ then (\ref{sdfsdf}) is an equality so we
get 
\begin{equation}
\label{contsfens0}
E(\| X \|^2) = \sum_{m=0}^{\m-1} \| E(\X_{m}) \|^2 + E(\|\X_{\m}\|^2) ~.
\end{equation}
To prove \eqref{nsdfoins}, we must show that $E(\|\X_{\m}\|^2)$ tends to $0$ when $\m$ goes to $\infty$. For all $M>0$,
\begin{align}
E(\|\X_{\m}\|^2)&= E(\|\X_{\m}\|^21_{\|X\|> M})+E(\|\X_{\m}\|^21_{\|X\|\leq M})\nonumber\\
&\leq E(\|\X_{\m}\|^21_{\|X\|> M})+ME(\|\X_{\m}\|1_{\|X\|\leq M}).
\label{division}
\end{align}
From \eqref{contsfens0}, $\| E(\X_{\m})\|\to 0$ when $\m\to\infty$. 
From the following lemma we shall derive that 
$E(\|\X_{\m}\|^2)  \to 0$ when $\m\to\infty$. 

\begin{lemma}\label{convergence}
$\lim_{\m\to\infty} \E(\|\X_{\m}\|1_{\|X\|\leq M}) = 0$.
\end{lemma}

Inserting the lemma result in \eqref{division} proves that
\begin{equation*}
\underset{\m}\limsup\, E(\|\X_{\m}\|^2)
\leq \underset{\m}{\limsup}\, E(\|\X_{\m}\|^21_{\|X\|> M})
\end{equation*}

Observe that
\[
\|\X_{\m} \|^2\leq \|\X_1\|^2+K ~~\mbox{with}~~K = \underset{m=1}{\overset{\infty}{\sum}}\|E(\X_m)\|^2,
\]
where $K < \infty$ because of \eqref{contsfens0}.
Indeed, $\X_{m}$ has positive coordinates so
\[
\|\X_{m+1}\|^2 =\|\X_{m}-E(\X_m)\|^2 \leq\|\X_m\|^2+\|E(\X_m)\|^2
\]
and hence
\[
\|\X_{\m}\|^2 \leq \|\X_1\|^2+\underset{m=1}{\overset{\m-1}{\sum}}\|E(\X_m)\|^2
\leq \|\X_1\|^2+\underset{m=1}{\overset{\infty}{\sum}}\|E(\X_m)\|^2.
\]

It results that for all $M>0$
\begin{equation*}
\underset{\m}\limsup\, E(\|\X_{\m}\|^2)
\leq \underset{\m}\limsup\, \Big(E(\|\X_1\|^21_{\|X\|> M})+KE(1_{\|X\|> M})\Big)~.
\end{equation*}
Since $\E(\|\X_1\|^2) \leq \E(\|X\|^2)<\infty$, 
the above limit tends to $0$ when $M\to\infty$ so $E(\|\X_{\m}\|^2)\to 0$.

Let us now prove Lemma \ref{convergence}.
Let $\epsilon$ be a positive number. Let $(E_1,...,E_T)$ be a partition of $\{x\in \R^{N_0},\|x\|_2\leq M\}$ in measurable non-empty sets such that, for all $t\leq T$, the diameter of $E_t$ is less than $\epsilon$.
For all $t\leq T$, we fix $x_t\in E_t$.
If $v_1,...,v_T\in\R^{N_{\m}}$ have positive coordinates then
\begin{equation*}
\underset{t\leq T}{\sum}\|v_t\|\leq \sqrt{T}
\left(\underset{t\leq T}{\sum}\|v_t\|^2 \right)^{1/2}
\leq \sqrt{T}\Big\|\underset{t\leq T}{\sum}v_t\Big\|.
\end{equation*}
We write $\X_m= \U_m X$.
Since $\U_0=Id$ and
$U_{m+1}x = |W_m (U _m x-E(\X_m))|$, each $\U_m$
is Lipschitz with constant $1$. It results that
$|\|U_m X \|- \|U_m x_t \|| \leq \|x_t - X\|$ and hence
\begin{align*}
E(\|\X_{\m}\|1_{\|X\|\leq M})
&=\underset{t\leq T}{\sum}E(\|\X_{\m}\|1_{X\in E_t})
=\underset{t\leq T}{\sum}E(\|U_{\m} X\|1_{X\in E_t})\\
&\leq\underset{t\leq T}{\sum}\Big(E(\|U_{\m} x_t\|1_{X\in E_t})+E(\|x_t-X\|1_{X\in E_t})\Big)\\
&=\underset{t\leq T}{\sum}
\Big(\|E(U_{\m} x_t\,1_{X\in E_t})\|+E(\|x_t-X\|1_{X\in E_t})\Big)\\
&\leq\underset{t\leq T}{\sum}\Big(\|E(U_{\m}X \,1_{X\in E_t})\|+2E(\|x_t-X\|1_{X\in E_t})\Big)\\
&\leq\sqrt{T}\Big\|\underset{t\leq T}{\sum}E(U_{\m} X\,1_{X\in E_t})\Big\|+2\epsilon\underset{t\leq T}\sum E(1_{X\in E_t})\\
&=\sqrt{T}\| E(\U_m X\,1_{\|X\|\leq M})\|+2\epsilon
\leq\sqrt{T}\|E(\X_{\m})\|+2\epsilon\\
\end{align*}
because $\X_{\m}$ has positive coordinates. It results from
\eqref{contsfens0} that $\lim_{\m \to \infty} \|E(\X_{\m})\|= 0$ so
\begin{equation*}
\underset{\m}{\limsup}\,E(\|\X_{\m}\|1_{\|X\|\leq M})
\leq 2\epsilon~.
\end{equation*}
Letting $\epsilon$ go to $0$ proves that
$\lim_{\m to \infty} E(\|\X_{\m}\|1_{\|X\|\leq M})= 0$.

Finally, we prove \eqref{nsdfoins2} by contradiction: we assume that $\underset{m=0}{\overset{\infty}{\sum}}\|E(\X_m)\|<\infty$.
\begin{gather*}
\forall m,\quad\quad
\|\X_{m+1}\|=\|\X_m-E(\X_m)\| \geq \|\X_m\|-\|E(\X_m)\|\\
\Rightarrow\quad
\forall \m,\quad\quad
\|\X_{\m}\|\geq\max(0,\|X\|-\underset{m=0}{\overset{\infty}{\sum}}\|E(\X_m)\|)
\end{gather*}
If $X$ is not bounded, this contradicts lemma \ref{convergence}: for $M$ large enough,
\begin{equation*}
\forall \m,\quad
E(\|\X_{\m}\|1_{\|X\|\leq M})\geq E(1_{\|X\|\leq M}\max(0,\|X\|-\underset{m=0}{\overset{\infty}{\sum}}\|E(\X_m)\|))>0
\end{equation*}

\section{Proof of Theorem \protect\ref{condsfth}}

Since $A$ is an unitary projector and $|W_m|$ is contractive
\begin{eqnarray}
\label{sdfsdf9}
\|\tx_m - \ty_m\|^2 &\leq&
\|\tx_{m-1} - \ty_{m-1} - A_{m-1} (\tx_{m-1} - \ty_{m-1})\|^2 \\
\nonumber
&=&\|\tx_{m-1} - \ty_{m-1}\|^2  - \|A_{m-1} \tx_{m-1} - A_{m-1} \ty_{m-1}) \|^2 .
\end{eqnarray}
It results that
\[
\|x - y\|^2 \geq \sum_{m=1}^{\m} 
\|A_{m-1} (\tx_{m-1} - \ty_{m-1}) \|^2 + \|\tx_{\m} - \ty_{\m}\|^2 ~.
\]
Letting $\m$ go to $\infty$ proves (\ref{contsf9}).
If we set $y = 0$ then (\ref{sdfsdf9}) is an equality so we
get 
\begin{equation}
\label{contsfens}
\| x \|^2 = \sum_{m=0}^{\m-1} \|A_m \tx_{m} \|^2 + \|\tx_{\m}\|^2 ~.
\end{equation}
Since the $A_m$ perform averages over blocks of size at most $M$,
and $\tx_m$ has positive coordinates
$\| A_m \tx_m \|^2 \geq \|  \tx_m \|^2 / M$. Since $W_m$ is unitary,
(\ref{intsdf22}) implies that
\[
\|\tx_{m+1} \|^2 = \|\tx_m - A_{m} \tx_m\|^2 = \|\tx_m\|^2 - \|A_{m} \tx_m\|^2 
\leq \|\tx_m\|^2 (1 - M^{-1})~.
\]
It implies that $\|x_\m \|^2 \leq \|x\|^2\, (1-M^{-1})^{\m}$.
Inserting this
in (\ref{contsfens}) gives (\ref{nsdfoins3}). 

\section{Proof of Proposition \protect\ref{ensdfonfops}}
Let us compute
\begin{eqnarray}
\nonumber
\|A_{m} \tX_{m} - E(\X_{m})\| &=&
\|A_{m} \tX_{m} - A_m \X_{m} + A_m \X_{m} - E(\X_{m})\| \\
\nonumber
&\leq&
\|A_m (\tX_{m} - \X_{m})\| + \| A_m \X_{m} - E(\X_{m})\| \\
\label{lsdfn8dsfs}
&\leq&
\|\tX_{m} - \X_{m}\| + \| A_m \X_{m} - E(\X_{m})\| 
\end{eqnarray}
But
\begin{eqnarray*}
\|\tX_{m} - \X_{m}\| &\leq &\||W_{m} (\tX_{m-1} - A_{m-1} \tX_{m-1})|
 - |W_{m} (\X_{m-1} - E(\X_{m-1}))| \| \\
&\leq &\||W_{m} (\tX_{m-1} - \X_{m-1} - (A_{m-1} \tX_{m-1} - A_{m-1} \X_{m-1})
- A_{m-1} \X_{m-1} +  E(\X_{m-1}))| \| \\
&\leq &\|W_{m} (Id - A_{m-1})(\tX_{m-1} - \X_{m-1}) \| + \| A_{m-1} \X_{m-1}) -
E(\X_{m-1})) \| \\
&\leq &\|\tX_{m-1} - \X_{m-1}\| + \|A_{m-1} \X_{m-1} -
E(\X_{m-1}) \| ~.
\end{eqnarray*}
Since $\tX_0 = \X_0$,
inserting iterativey this equation in the previous equation 
together with (\ref{lsdfn8dsfs}) proves that
\begin{equation}
\label{dsfoins}
\|A_{m} \tX _{m} - E(\X_{m})\| \leq \sum_{n=0}^m \|A_n \X_{n} - E(\X_{n})\| ~.
\end{equation}
Since $E \left(\sum_n Y_n \right)^2 \leq
 \left( \sum_{n} E(|Y_n|^2)^{1/2} \right)^2$, we derive
(\ref{dsfoinsd3}) from (\ref{dsfoins}).

\section{Proof of Theorem \protect\ref{firnsdfpors}}
\label{blabla2} 

This theorem is proved by applying Proposition \ref{ensdfonfops}.
To prove (\ref{dsfoinsd5}), we 
define an aggregated random vector $Y = \{X_i \}_{i \leq P} \in \R^{N \times P}$.
The expected scattering tranform 
of $Y$ is $E(\Y_m)= \{E(\X_{i,m} ) \}_{i \leq P}$ and since each
$X_i$ has the same distribution as $X$ it results that
$E(\X_{i,m}) = E(\X_{m})$. 
Observe that $\tY_m = \{\tX_{i,m} \}_{i \leq P}$ is an averaged scattering
transform computed with
\begin{equation}
\label{intsdf223}
{\tY}_{m+1} = |W_{m+1} (\tY_{m} - A_{m} \tY_{m})|
\end{equation}
where $A_m \tY_{m}$ is a vector of size $P N_m$ and is a concatenation
$P$ vectors equal to $\bar \mu_m = P^{-1} \sum_{i=1}^P \tX_{i,m}$.
The vector $E(\Y_{m})$ is also of size $P N_m$ and is a  concatenation
of $P$ vectors equal to $E(\X_m)$.
Applying Proposition \ref{dsfoinsd3} to $\tY_m$ proves that
\begin{equation}
\label{nsdfosdf}
E(\|A_m \tY_{m} - E(\Y_{m})\|^2) \leq 
\left(\sum_{n=0}^m E(\|A_n \Y_{n} - E(\Y_{n})\|^2)^{1/2} \right)^2 ~.
\end{equation}
But $E(\|A_m \tY_{m} - E(\Y_{m})\|^2) = P E(\|\bar \mu_m - E(\X_m) \|^2)$.
Moreover $\Y_n = \{\X_{i,n} \}_{i \leq P}$ is a concatenation of $P$
independent random vectors of same distribution as $X_n$ so,
\[
E(\|A_n \Y_{n} - E(\Y_{n})\|^2 =  E(\|\X_{n} - E(\X_{n})\|^2 = 
 \sum_{k=n}^{+\infty} E(\| \X_{k}\|^2)~.
\]
Inserting these two equations in (\ref{nsdfosdf}) 
together with $E(\|\X_k \|^2) \leq
E(\|X\|^2)$ proves the two inequalities of (\ref{dsfoinsd5}).

\bibliographystyle{plain}

\end{document}